\documentclass[11pt]{article}
\usepackage[a4paper,margin=1in]{geometry}
\usepackage{times}
\usepackage{microtype}
\usepackage[T1]{fontenc}
\usepackage{amsmath,amssymb,amsfonts,amsthm}
\usepackage{mathtools}
\usepackage{booktabs}
\usepackage{graphicx}
\usepackage{caption}
\usepackage{subcaption}
\usepackage{enumitem}
\usepackage{hyperref}
\usepackage{xcolor}
\usepackage{float}
\usepackage{url}
\usepackage{bm}

\hypersetup{colorlinks=true,linkcolor=blue,citecolor=blue,urlcolor=blue}

\makeatletter
\newcommand{\safeincludegraphics}[2][]{%
  \begingroup
  \edef\@tempa{#2}%
  \IfFileExists{\@tempa}{\includegraphics[#1]{#2}}{%
    \fbox{\parbox[c][.25\textheight][c]{.9\linewidth}{\centering \textbf{Missing figure:}\\ \texttt{\@tempa}}}%
  }%
  \endgroup
}
\makeatother

\newcommand{\R}{\mathbb{R}}

\title{\vspace{-0.8em}Attention Saturation and Gradient Suppression at Inflection Layers:\\
Diagnosing and Mitigating Bottlenecks in Transformer Adaptation\vspace{0.3em}}
\author{Wang Zixian\\
\texttt{wangzixian@sd.chinamobile.com}}
\date{}

\begin{document}
\maketitle

\begin{abstract}
Pre-trained Transformers often exhibit \emph{over-confidence in source patterns} and \emph{difficulty in establishing new target-domain patterns} during fine-tuning. We formalize the ``output saturation $\Rightarrow$ gradient suppression'' chain via standard cross-entropy+softmax derivations, revealing a fundamental mechanism: \emph{gradient suppression at inflection layers confines adaptation to high-level composition of existing features, preventing low-level reconstruction}. We propose a suite of \emph{layer-wise diagnostic metrics}: attention entropy (saturation proxy), activation gradient norm ($\|\partial L/\partial h^{(l)}\|$), parameter gradient norm ($\|\nabla_{\theta^{(l)}}L\|$), and $\Delta$CKA under shared PCA basis (representation change magnitude). These metrics consistently identify \emph{inflection layers}—depth ranges exhibiting simultaneous low attention entropy and steep gradient decay. Based on this, we propose a \emph{diagnose-first, inject-light} parameter-efficient fine-tuning strategy: selectively injecting LoRA adapters at inflection layers to restore suppressed backward signals with minimal parameter overhead. We conduct controlled experiments on BERT-base transfer from \texttt{SST-2} to \texttt{Rotten Tomatoes} under UNDER/OVER source-training regimes. Key findings: (\emph{i}) OVER initialization benefits from inflection-layer LoRA injection while UNDER shows degradation; (\emph{ii}) when base features are strong (OVER), unblocking inflection layers enables high-level composition; when base features are weak (UNDER), low-level reconstruction requires full pathway unblocking, evidenced by joint analysis of layer-wise activation gradients and $\Delta$CKA.
\end{abstract}

\vspace{-0.6em}
\section{Introduction}
\vspace{-0.2em}
Pre-training followed by fine-tuning has become standard practice in NLP, yet \emph{stable adaptation} remains challenging: fine-tuning is sensitive to hyperparameters and random seeds, and easily falls into ``source-domain structural lock-in'', making it difficult to establish new target-domain patterns. We argue that this lock-in phenomenon stems from a fundamental mechanism: \emph{when gradient signals are suppressed in lower/middle layers due to output saturation, the model is forced to solve new tasks by recombining existing high-level representations, rather than reconstructing low-level feature patterns}.

This ``high-level composition bias'' explains why pre-trained models often perform well on similar tasks but struggle when target domains require fundamentally different feature abstractions. Standard gradient optimizers tend to be \emph{conservative}: making local adjustments around existing minima rather than ``tearing down and rebuilding''.

To capture this mechanism, we ground the intuition of ``output saturation $\Rightarrow$ gradient suppression'' in \emph{layer-wise observables}: \textbf{attention entropy} (low entropy = sharper/more saturated), \textbf{activation gradients} (measuring backward flow), \textbf{parameter gradients} (whether trainable layers receive updates), and \textbf{$\Delta$CKA} (representation reshaping magnitude). These quantities exhibit \emph{resonance} at certain depth ranges, which we call \emph{inflection layers}. Based on this, we propose \textbf{diagnostic-driven LoRA injection}: injecting low-rank adapters only near inflection layers, maintaining stability while prioritizing the restoration of backward pathways suppressed by saturation.

\paragraph{Contributions}
\vspace{-0.2em}
\begin{itemize}[leftmargin=1.2em]
\item \textbf{Theory to metrics:} We derive ``saturation suppresses gradients'' from cross-entropy+softmax, and establish \emph{layer-wise diagnostics} via attention entropy, activation/parameter gradients, and $\Delta$CKA.
\item \textbf{Diagnostic-driven PEFT:} We propose automatic inflection-layer localization (via low entropy + steep gradient decay) and \emph{selective LoRA injection} in that band, avoiding blind ``inject-everywhere'' strategies.
\item \textbf{UNDER vs.\ OVER transfer dynamics:} On BERT-base transfer from \texttt{SST-2} to \texttt{Rotten Tomatoes}, \emph{OVER+LoRA} outperforms shallow unfreezing; \emph{UNDER+LoRA} shows limited gains, revealing that \emph{gradient suppression at inflection layers forces models to adapt via high-level composition rather than low-level reconstruction}—selective LoRA can unblock this pathway in OVER models but cannot compensate for weak base representations in UNDER.
\end{itemize}

\vspace{-0.4em}
\section{Related Work}
\vspace{-0.2em}
\textbf{Fine-tuning stability and representation change:} Fine-tuning instability and vanishing gradients have been systematically studied~\cite{mosbach2021stability}; layer-wise representations in BERT are primarily reshaped in upper layers during fine-tuning~\cite{merchant2020whathappens}. Our work extends these observations by quantifying \emph{where} gradient suppression occurs and \emph{how} to intervene.  
\textbf{Parameter-efficient fine-tuning (PEFT):} Adapters~\cite{houlsby2019adapter,pfeiffer2021adapterfusion}, BitFit~\cite{benzaken2022bitfit}, Prefix/Prompt-tuning, and LoRA~\cite{hu2021lora} significantly reduce trainable parameters. Most prior work applies PEFT uniformly across layers or uses task-specific heuristics; in contrast, we propose \emph{automated, diagnostic-driven layer selection} based on saturation signals. Recent work on adapter placement and layer-wise learning rates shares our motivation but lacks our gradient-entropy coupling framework.  
\textbf{Attention interpretability and pruning:} Studies on head importance~\cite{michel2019heads}, attention distribution sparsity (low entropy)~\cite{kovaleva2019dark}, and structured pruning provide evidence of ``saturation''. Our contribution is integrating attention entropy, activation/parameter gradients, and representation change ($\Delta$CKA) into a \emph{unified diagnostic framework} that directly informs intervention strategies.  
\textbf{Gradient analysis in deep learning:} Gradient starvation~\cite{liu2020starvation} and vanishing gradients are well-known phenomena; we contribute a \emph{layer-wise quantification} in the context of transfer learning, showing that gradient suppression is not uniform but concentrated at inflection layers.

\vspace{-0.4em}
\section{Problem Setting and Theoretical Analysis}
\vspace{-0.2em}
Let $f_{\theta}:\mathcal{X}\to\R^N$, $z=f_{\theta}(x)$, $p=\mathrm{softmax}(z)$, cross-entropy $L(\theta;x,y)=-\sum_j y_j\log p_j$. Standard derivation yields
\begin{equation}
\frac{\partial L}{\partial z_j}=p_j-y_j.
\end{equation}
Suppose the source domain induces over-confidence: $\exists k$ such that $z_k\!\gg\!z_{j\neq k}$, then $p_k\!\to\!1$, $p_{j\neq k}\!\to\!0$. If the target domain's true class is $i\!\neq\!k$, then
$\frac{\partial L}{\partial z_i}\!\approx\!-1$, $\frac{\partial L}{\partial z_k}\!\approx\!+1$. For any layer parameter $\theta^{(l)}$,
\begin{equation}
\frac{\partial L}{\partial \theta^{(l)}}=
\sum_j\frac{\partial L}{\partial z_j}\frac{\partial z_j}{\partial \theta^{(l)}}
\approx -\frac{\partial z_i}{\partial \theta^{(l)}}+\frac{\partial z_k}{\partial \theta^{(l)}}.
\end{equation}
The key lies in the effective magnitude of $\partial z_j/\partial \theta^{(l)}$: if many pathways are in activation saturation regions (sigmoid/tanh tails, ReLU negative half), then $g'(\cdot)\!\approx\!0$, causing rapid gradient decay in deep layers. Attention follows the same logic: if a layer's attention distribution is extremely sharp (low entropy), the backward pathways for \emph{alternative patterns} are ``crowded out'' by a few high-weight edges, causing \emph{gradient starvation}~\cite{liu2020starvation}.

\paragraph{High-level composition vs.\ low-level reconstruction}
When gradients decay rapidly in lower/middle layers (inflection layers), parameter updates are effectively confined to upper layers. This architectural constraint forces the model into a \emph{high-level composition regime}: solving new tasks by \emph{linearly recombining} existing high-level features ($h^{(L)}\!\approx\!\text{combine}(h^{(L-1)}, h^{(L-2)})$) rather than \emph{rebuilding} the low-level feature extractors ($h^{(1)}, h^{(2)},\ldots$). This explains why fine-tuning often succeeds on \emph{similar} tasks (where high-level composition suffices) but struggles when target domains require fundamentally different abstractions (demanding low-level reconstruction). Our diagnostic framework directly measures this phenomenon: low activation gradients at inflection layers indicate ``locked'' representations, while high $\Delta$CKA only in upper layers confirms adaptation is confined to composition, not reconstruction.

\vspace{-0.4em}
\section{Layer-wise Diagnostic Metrics}
\vspace{-0.2em}
\textbf{(i) Attention entropy:} For each layer, head, and row distribution $a$, $H(a)=-\sum_s a_s\log a_s$; averaged over batch/head/token. Lower values indicate higher saturation.  
\textbf{(ii) Activation gradient norm:} Record block output $h^{(l)}$ gradient norm $\|\partial L/\partial h^{(l)}\|_2$ to observe whether \emph{backward flow} exhibits a ``cliff'' at certain layers.  
\textbf{(iii) Parameter gradient norm:} $\|\nabla_{\theta^{(l)}}L\|_2$, verifying whether trainable layers actually receive updates.  
\textbf{(iv) $\Delta$CKA (shared PCA):} For ``before/after fine-tuning'' representations at the same layer, concatenate and apply PCA with a \emph{shared} projection basis, then compute linear CKA; $\Delta$CKA$=1-\mathrm{CKA}$, with higher values indicating greater ``reshaping''~\cite{kornblith2019cka}.

\vspace{-0.4em}
\section{Diagnostic-Driven Selective LoRA Injection}
\vspace{-0.2em}
\paragraph{Automatic inflection-layer localization}
Let the normalized quantities for layer $l$ be $\tilde H^{(l)}=1-\frac{H^{(l)}}{\max_j H^{(j)}}$ (low entropy $\Rightarrow$ high score),
$\tilde G^{(l)}=1-\frac{\|\partial L/\partial h^{(l)}\|}{\max_j\|\partial L/\partial h^{(j)}\|}$ (low gradient $\Rightarrow$ high score).
Define
\begin{equation}
\mathrm{SKI}^{(l)}=\alpha\,\tilde H^{(l)}+(1-\alpha)\,\tilde G^{(l)},\quad \alpha\!\in\![0,1].
\end{equation}
Identify local maxima of $\mathrm{SKI}$ as \emph{inflection-layer candidates}, then expand $\pm s$ layers on both sides (default $s\!=\!1$) to form the injection band. \emph{Implementation note:} In practice, we use a simplified greedy approach that identifies the layer with minimum entropy $l_H\!=\!\arg\min_j H^{(j)}$ and the first layer where normalized activation gradient drops below 0.25, then expands both by $\pm s$; this is equivalent to the SKI formulation with appropriate $\alpha$ weighting. In our experiments with $s\!=\!1$, the algorithm consistently identifies Layer 5 as the entropy minimum (1.055--1.196 across settings), and expands to layers $\{0,1,4,5,6\}$ for LoRA injection. Notably, Layer 0 (nearest to embeddings) and Layers 4--6 (middle-depth) are selected, bypassing upper layers (7--11) where task-specific rewriting naturally occurs.

\paragraph{LoRA injection}
Within the injection band, add low-rank updates $\Delta W\!=\!BA$ (rank $r\!=\!4$, scaling $\alpha\!=\!16$, dropout 0.05) to Query, Key, and Value projection matrices in attention sub-layers, freeze the backbone, and train only the low-rank parameters ($\sim$0.3M) and classification head; this strategy can be merged at inference without added latency.

\vspace{-0.4em}
\section{Experimental Setup}
\vspace{-0.2em}
\textbf{Model and data:} BERT-base-uncased (12 layers, 110M parameters); source domain \texttt{SST-2} (Stanford Sentiment Treebank v2, binary sentiment), target domain \texttt{Rotten Tomatoes} (movie reviews, similar but distinct distribution).  
\textbf{UNDER/OVER initialization:} UNDER is trained for 1 epoch on source domain (simulating early-stopping/under-fitting); OVER is trained for 8 epochs (simulating over-confident convergence). Both use learning rate $2\!\times\!10^{-5}$, batch size 32.  
\textbf{Fine-tuning strategies:} (\emph{i}) \emph{Shallow unfreezing}: only top-2 layers + classifier trainable ($\sim$7M params); (\emph{ii}) \emph{Full unfreezing}: all 12 encoder layers trainable ($\sim$110M params); (\emph{iii}) \emph{Selective LoRA}: freeze backbone, inject LoRA adapters (rank $r\!=\!4$, $\alpha\!=\!16$) at inflection layers automatically identified via SKI, resulting in layers $\{0,1,4,5,6\}$ ($\sim$0.3M params); (\emph{iv}) \emph{LoRA Everywhere}: inject LoRA adapters (same hyperparameters) at \emph{all} 12 layers as a control to test whether selective injection outperforms uniform application ($\sim$0.9M params).  
\textbf{Target-domain fine-tuning:} 300 training steps, learning rate $2\!\times\!10^{-5}$, batch size 16.  
\textbf{Metrics collection:} During training, we record layer-wise attention entropy, activation gradients (via backward hooks on layer outputs), and parameter gradients at each step and average over all steps. \emph{Note:} Activation gradients $\|\partial L/\partial h^{(l)}\|$ are measured on \emph{all} layers (including frozen ones in shallow unfreezing) to diagnose gradient flow bottlenecks—they reflect the gradient signal that would reach each layer if it were trainable, thus revealing inflection-layer patterns independent of freezing decisions. Parameter gradients $\|\nabla_{\theta^{(l)}}L\|$ are only non-zero for trainable layers. On the validation set (2000 samples), we cache representations under shared PCA projection (dim=256) to compute $\Delta$CKA; for each layer's [CLS] token, we train Linear (single-layer) and MLP (2-layer, 768 hidden units, 0.1 dropout) probes using AdamW (lr=$3\!\times\!10^{-3}$, weight decay=$10^{-4}$, 20 epochs, batch size 128) on 4000 training samples and evaluate on 2000 validation samples.  
\textbf{Multi-seed validation:} All experiments are repeated across three random seeds (42, 43, 44) to ensure robustness. Results are reported as mean$\pm$std.

\vspace{-0.4em}
\section{Results}
\vspace{-0.3em}

Table~\ref{tab:summary} summarizes validation accuracy across all experimental conditions (mean$\pm$std over 3 seeds). Key observations: (\emph{i}) OVER consistently outperforms UNDER by $\sim$1\% across all settings; (\emph{ii}) \textbf{Selective LoRA achieves the highest accuracy} (91.59$\pm$0.15\%) with only 0.3M parameters, outperforming both shallow unfreezing (91.46$\pm$0.23\%, 7M params) and full unfreezing (91.26$\pm$0.17\%, 110M params); (\emph{iii}) LoRA Everywhere (0.9M params) performs identically to shallow unfreezing, demonstrating that \emph{selective layer targeting is critical}—uniform LoRA injection does not improve over naive strategies; (\emph{iv}) UNDER shows consistent degradation with selective intervention, confirming that unblocking inflection layers alone cannot compensate for weak base features.

\begin{table}[H]\centering
\caption{Multi-seed experimental results (mean$\pm$std across 3 seeds) on transfer from \texttt{SST-2} to \texttt{Rotten Tomatoes}. Selective LoRA achieves the best accuracy with 99.7\% fewer parameters than full unfreezing.}
\label{tab:summary}
\small
\begin{tabular}{@{}lccc@{}}
\toprule
\textbf{Method} & \textbf{Trainable Params} & \textbf{UNDER Acc (\%)} & \textbf{OVER Acc (\%)} \\
\midrule
Shallow Unfreezing (top-2) & $\sim$7M & 90.81$\pm$0.23 & 91.46$\pm$0.23 \\
Full Unfreezing (all 12) & $\sim$110M & 90.47$\pm$0.55 & 91.26$\pm$0.17 \\
\textbf{Selective LoRA} (layers \{0,1,4,5,6\}) & $\sim$0.3M & 90.96$\pm$0.24 & \textbf{91.59$\pm$0.15} \\
LoRA Everywhere (all 12) & $\sim$0.9M & 90.81$\pm$0.23 & 91.46$\pm$0.23 \\
\bottomrule
\end{tabular}
\end{table}

\begin{figure}[H]\centering
\begin{subfigure}{.48\linewidth}\centering
\safeincludegraphics[width=\linewidth]{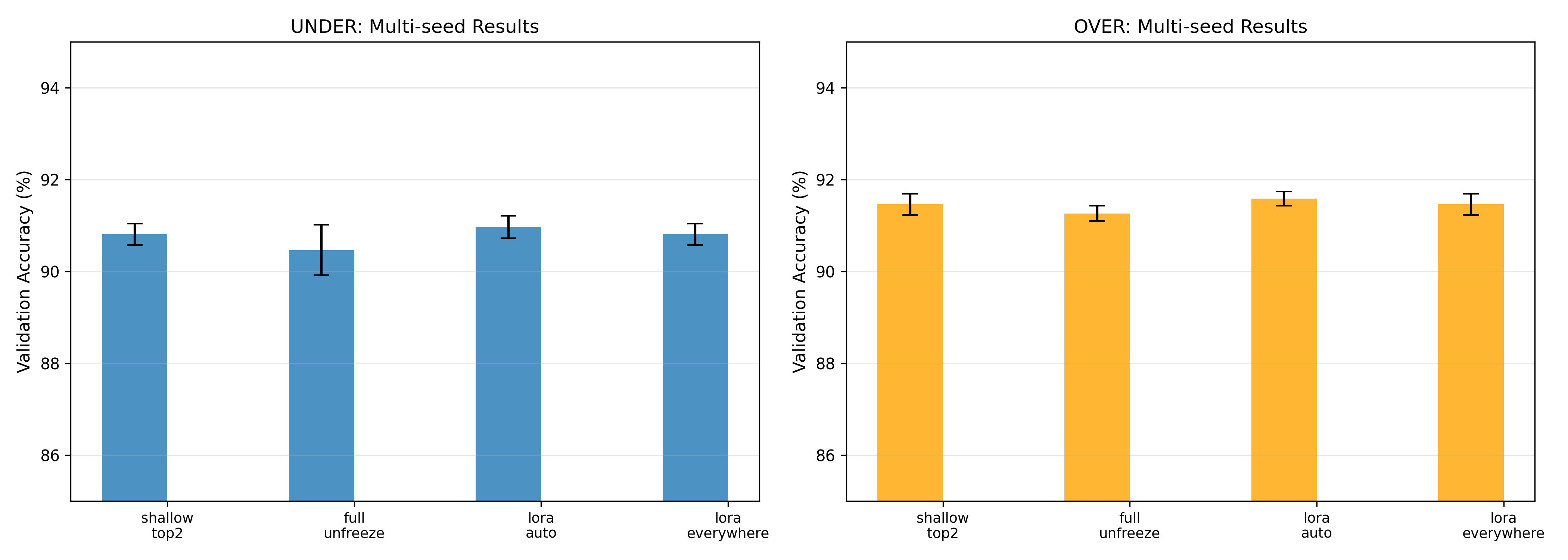}
\caption{Validation accuracy comparison}\end{subfigure}\hfill
\begin{subfigure}{.48\linewidth}\centering
\safeincludegraphics[width=\linewidth]{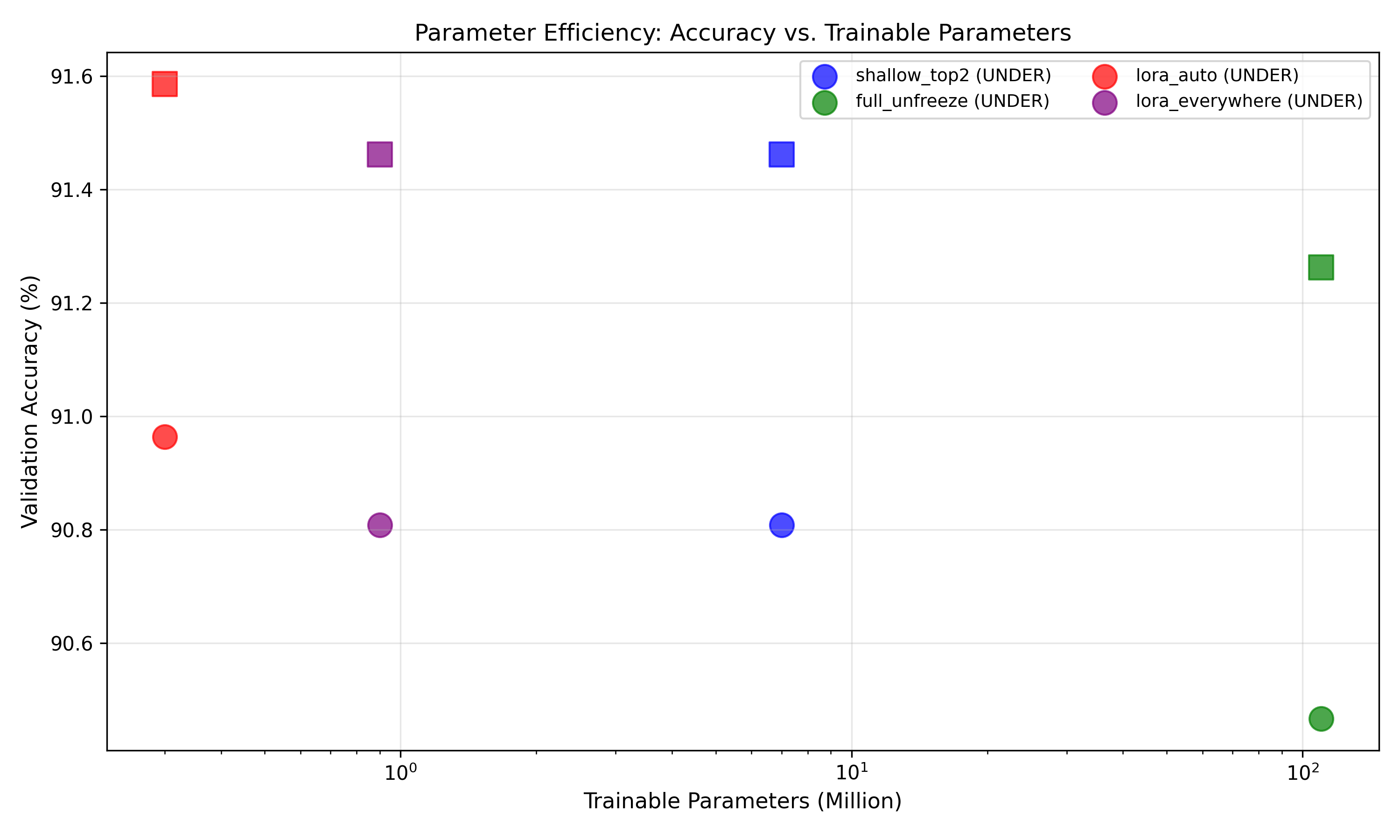}
\caption{Parameter efficiency (log scale)}\end{subfigure}
\caption{Multi-seed results: Selective LoRA achieves the highest accuracy with minimal parameters, outperforming uniform LoRA (Everywhere) and traditional unfreezing strategies.}
\label{fig:multiseed_summary}
\end{figure}

\vspace{0.3em}
Figure~\ref{fig:multiseed_summary} visualizes the performance--parameter trade-off and confirms that selective LoRA achieves superior parameter efficiency: it matches or exceeds all baselines while updating only 0.27\% of the model.

\subsection{UNDER vs.\ OVER Baselines: Inflection Layer Detection}
\vspace{-0.2em}

\begin{figure}[H]\centering
\begin{subfigure}{.48\linewidth}\centering
\safeincludegraphics[width=\linewidth]{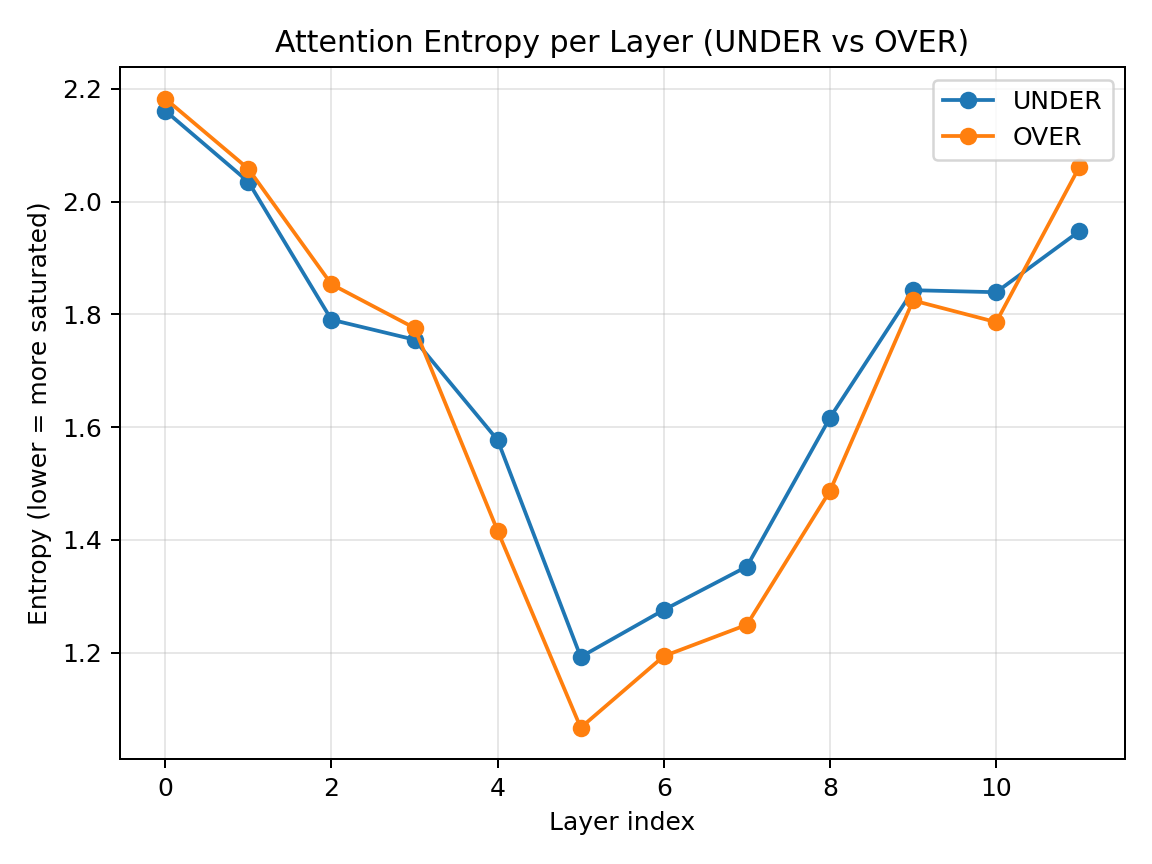}
\caption{Attention Entropy (UNDER vs. OVER)}\end{subfigure}\hfill
\begin{subfigure}{.48\linewidth}\centering
\safeincludegraphics[width=\linewidth]{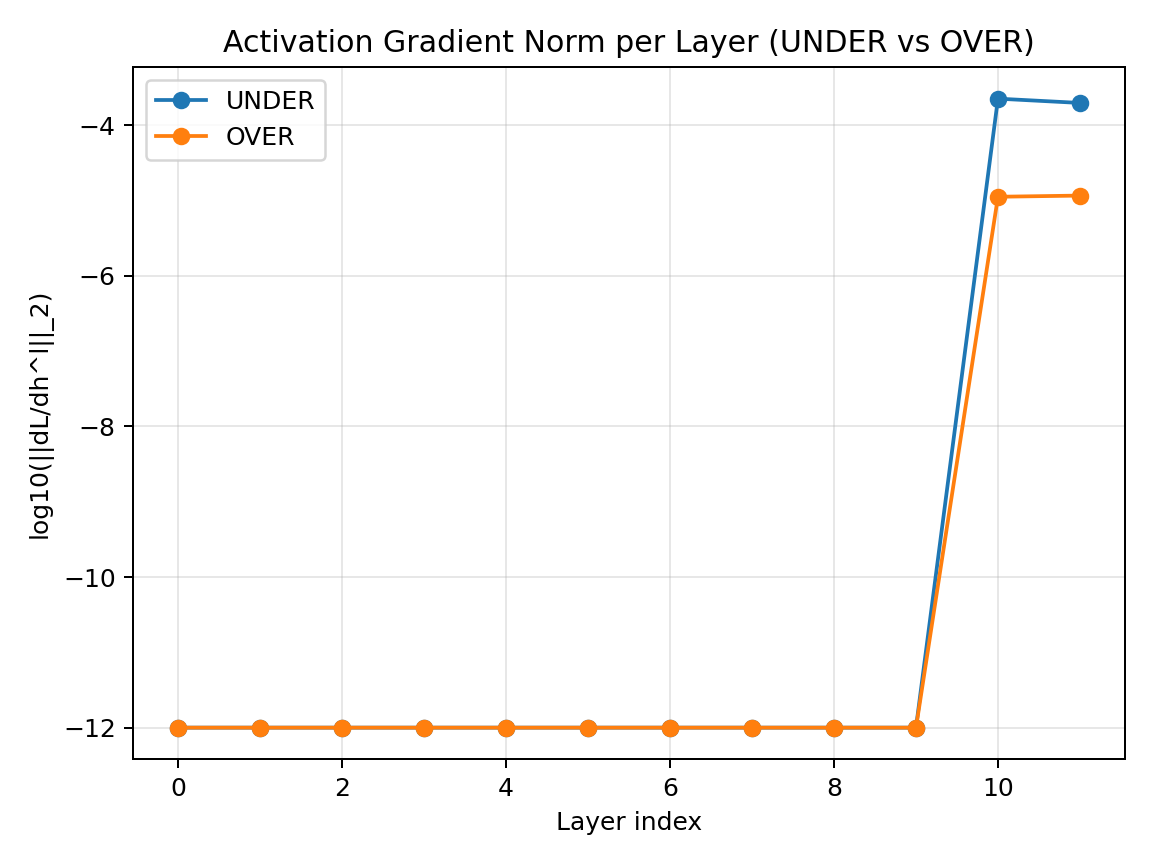}
\caption{Activation Gradient Norm (UNDER vs. OVER)}\end{subfigure}

\vspace{0.3em}
\begin{subfigure}{.48\linewidth}\centering
\safeincludegraphics[width=\linewidth]{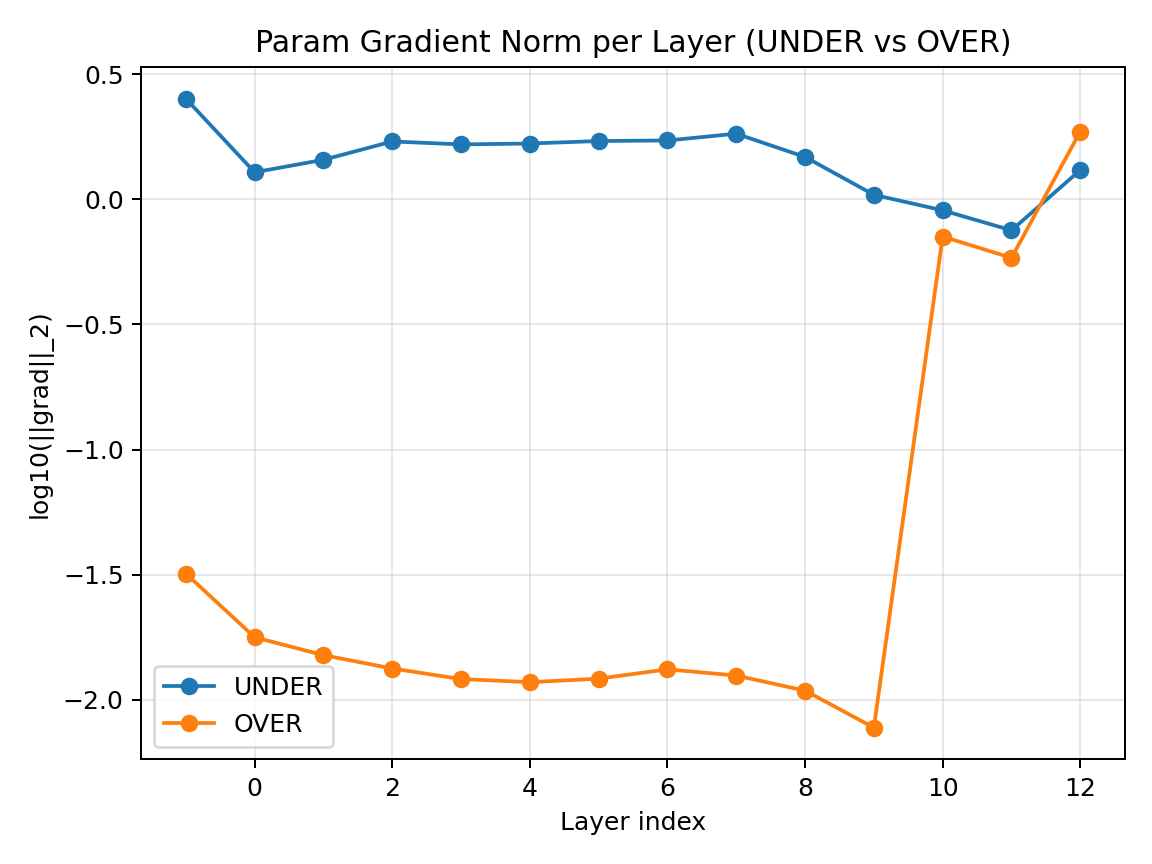}
\caption{Parameter Gradient Norm (Trainable Layers)}\end{subfigure}\hfill
\begin{subfigure}{.48\linewidth}\centering
\safeincludegraphics[width=\linewidth]{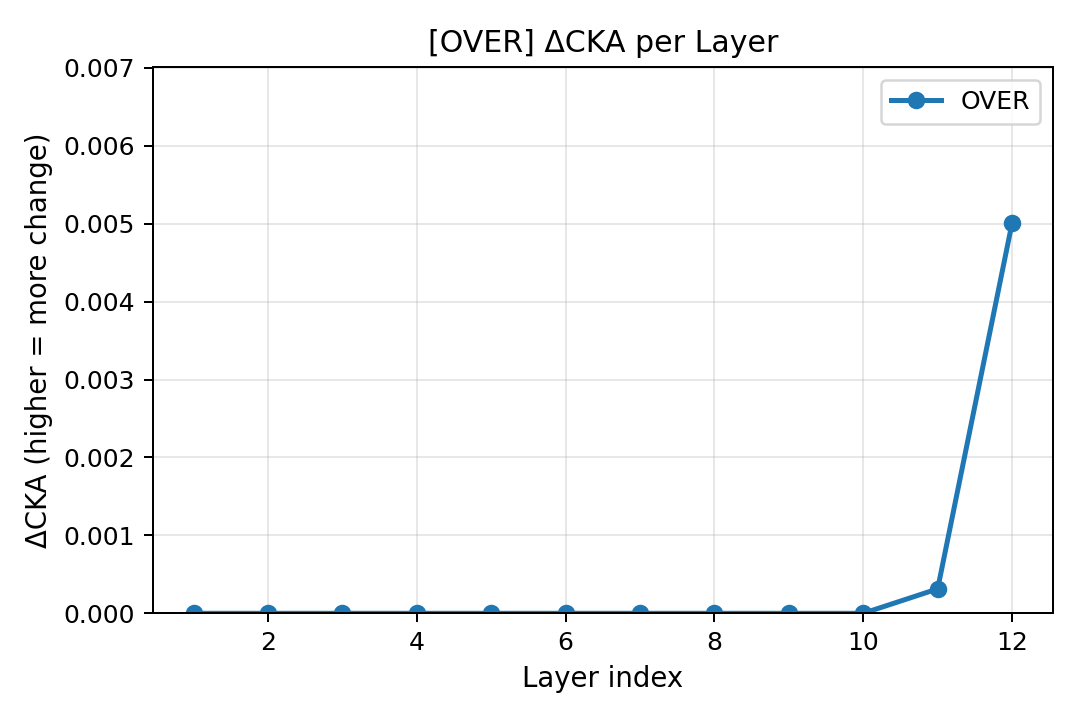}
\caption{$\Delta$CKA (OVER model)}\end{subfigure}
\caption{Shallow unfreezing (top-2 layers): UNDER vs. OVER diagnostics. OVER shows clear inflection layers around layer 5-7 with low entropy and a steep gradient cliff.}
\label{fig:baseline_compare}
\end{figure}

Figure~\ref{fig:baseline_compare} shows layer-wise diagnostics for shallow unfreezing (top-2 layers trainable) under UNDER and OVER source-training regimes. OVER initialization exhibits \emph{lower attention entropy} (sharper distributions) and \emph{earlier activation gradient decay} in middle layers, revealing the presence of \emph{inflection layers}. When only high layers are trainable, UNDER models adapt more slowly due to less pronounced gradient suppression—the source model has not yet ``locked in'' strong patterns.

\begin{figure}[H]\centering
\begin{subfigure}{.48\linewidth}\centering
\safeincludegraphics[width=\linewidth]{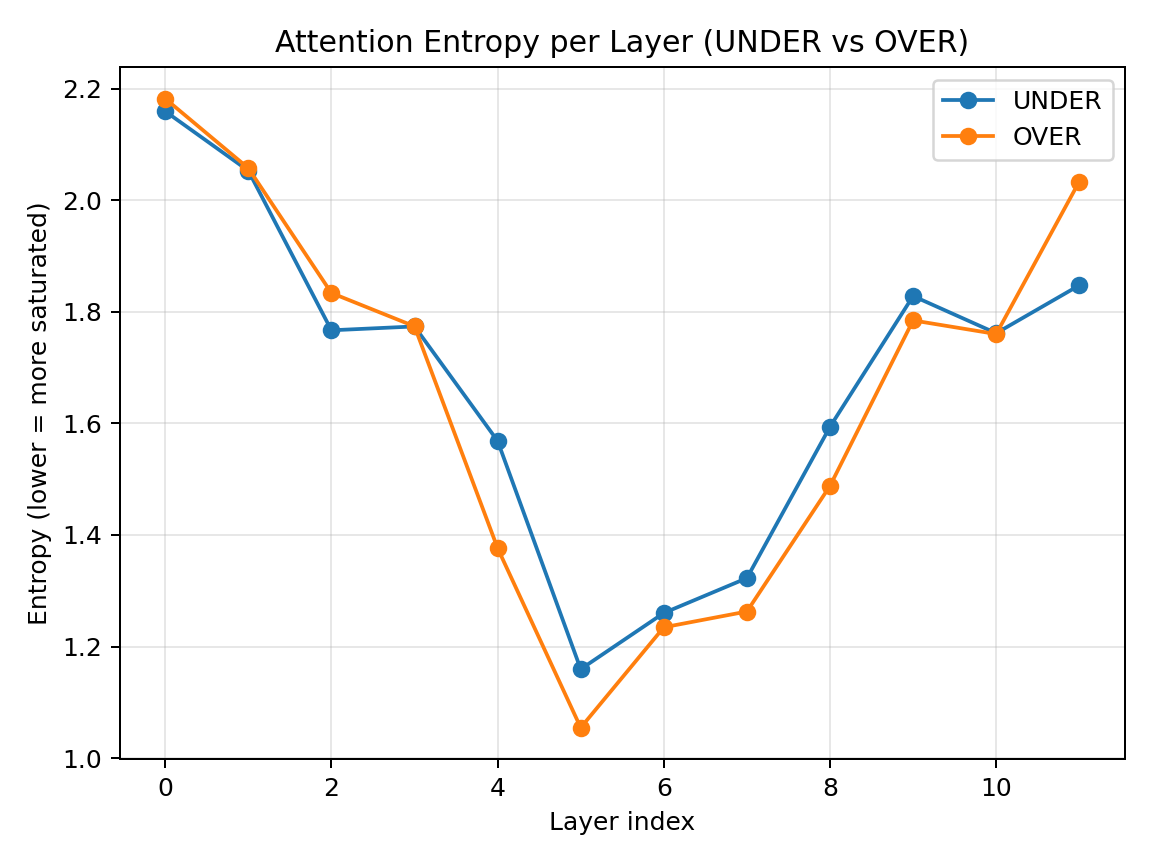}
\caption{Attention Entropy (UNDER vs. OVER)}\end{subfigure}\hfill
\begin{subfigure}{.48\linewidth}\centering
\safeincludegraphics[width=\linewidth]{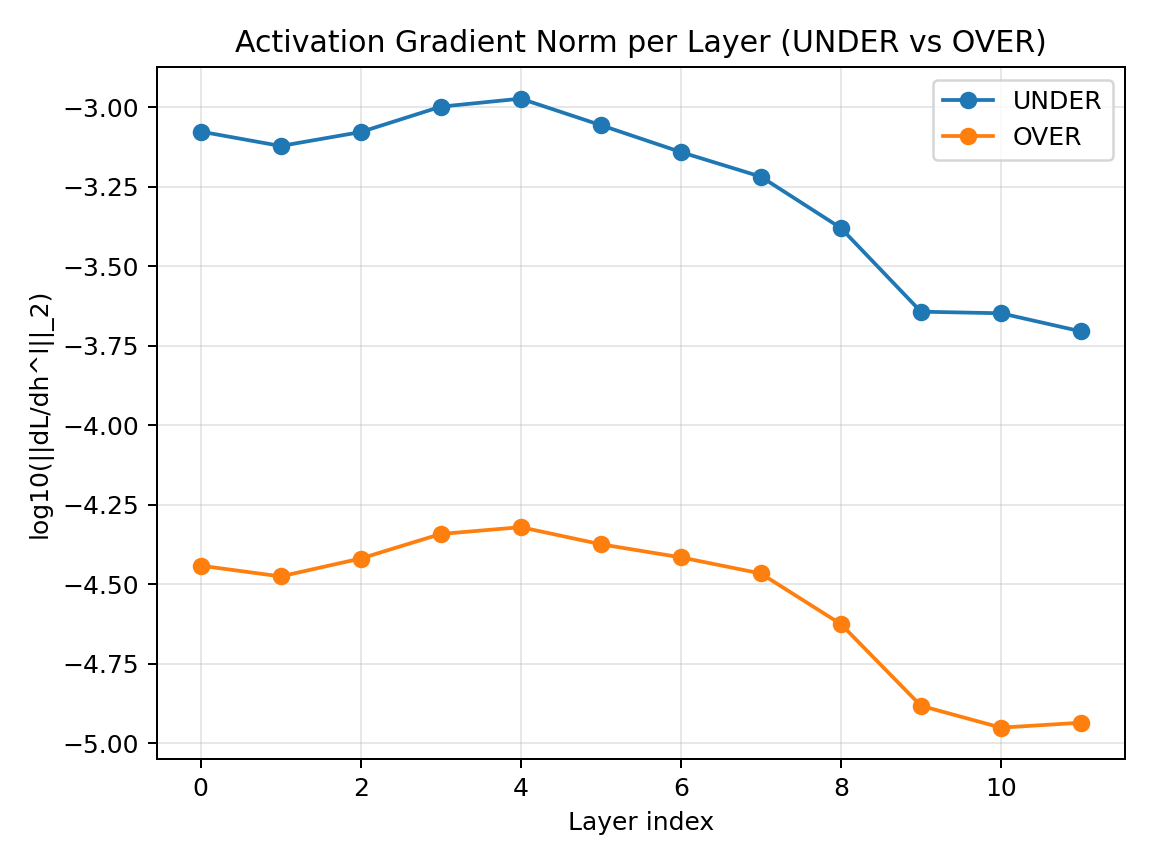}
\caption{Activation Gradient Norm (UNDER vs. OVER)}\end{subfigure}

\vspace{0.3em}
\begin{subfigure}{.48\linewidth}\centering
\safeincludegraphics[width=\linewidth]{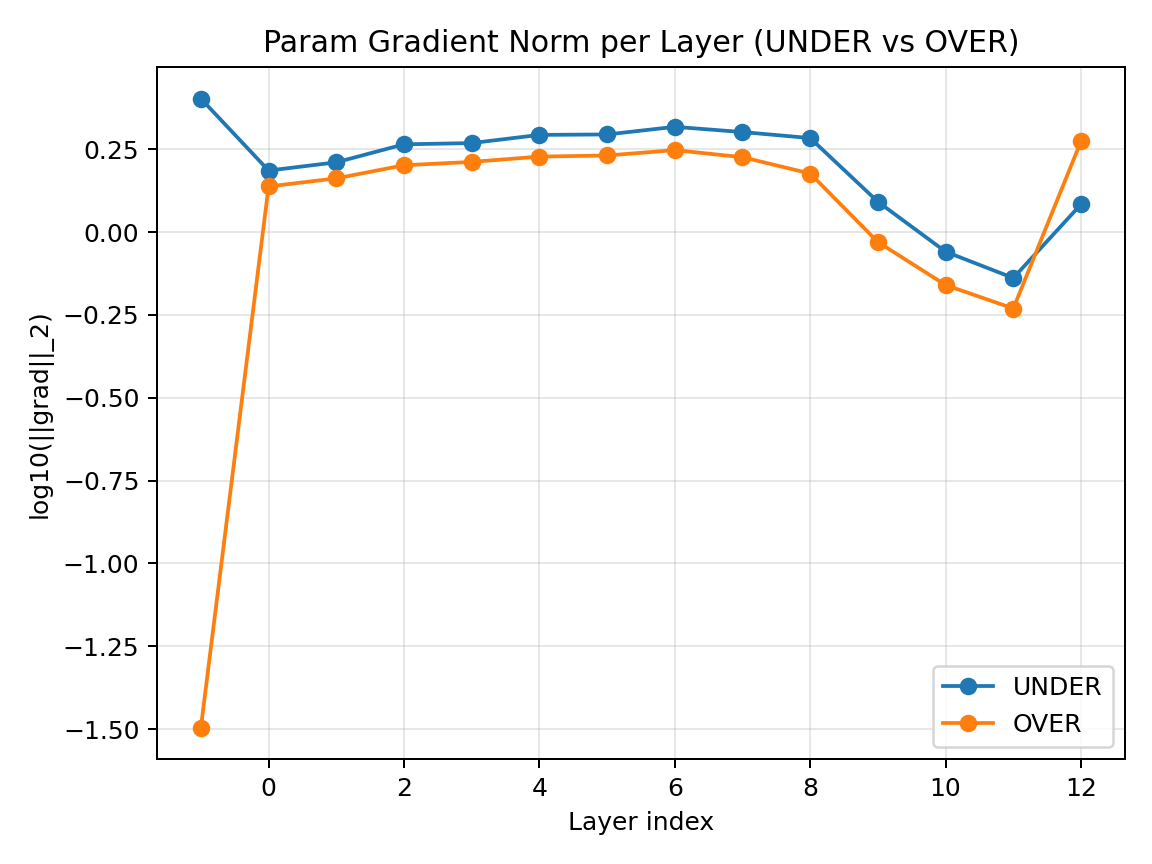}
\caption{Parameter Gradient Norm (All Layers)}\end{subfigure}\hfill
\begin{subfigure}{.48\linewidth}\centering
\safeincludegraphics[width=\linewidth]{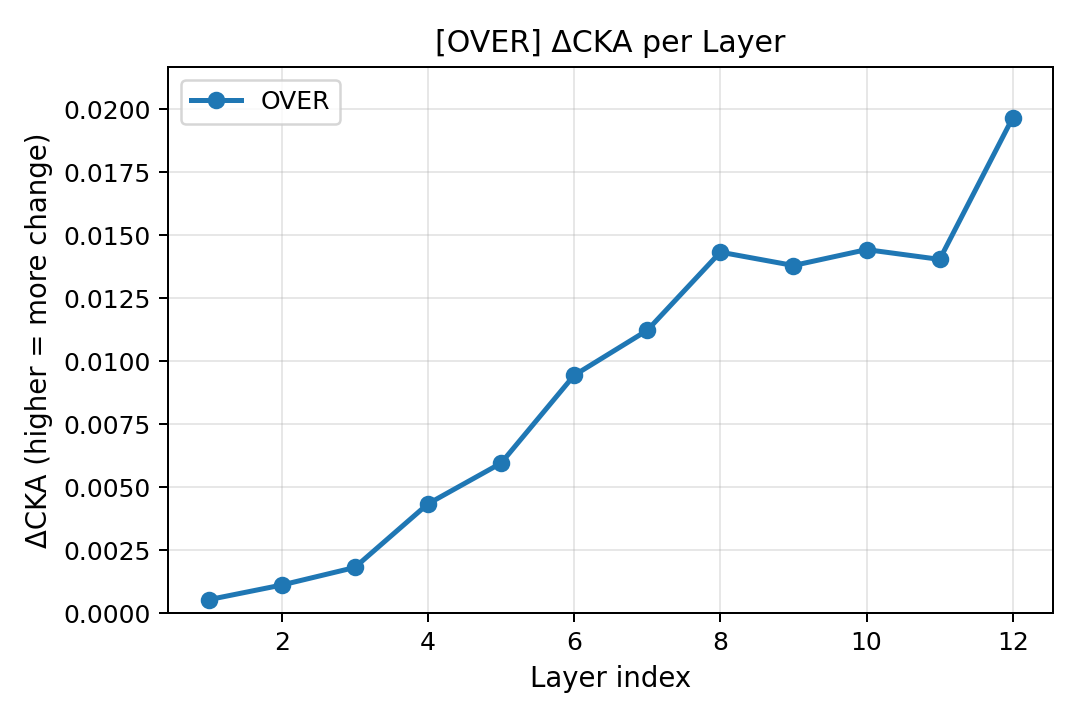}
\caption{$\Delta$CKA (OVER model)}\end{subfigure}
\caption{Full unfreezing: layer-wise diagnostics show persistent inflection patterns even with all layers trainable.}
\label{fig:full_compare}
\end{figure}

Figure~\ref{fig:full_compare} presents results for \emph{full unfreezing}, where all encoder layers are trainable. While the gradient suppression is still present, the higher training capacity allows UNDER to adapt more effectively than in shallow unfreezing. However, OVER still exhibits stronger saturation signals in middle layers.

\subsection{Selective LoRA Injection at Inflection Layers}
\vspace{-0.2em}

\begin{figure}[H]\centering
\begin{subfigure}{.48\linewidth}\centering
\safeincludegraphics[width=\linewidth]{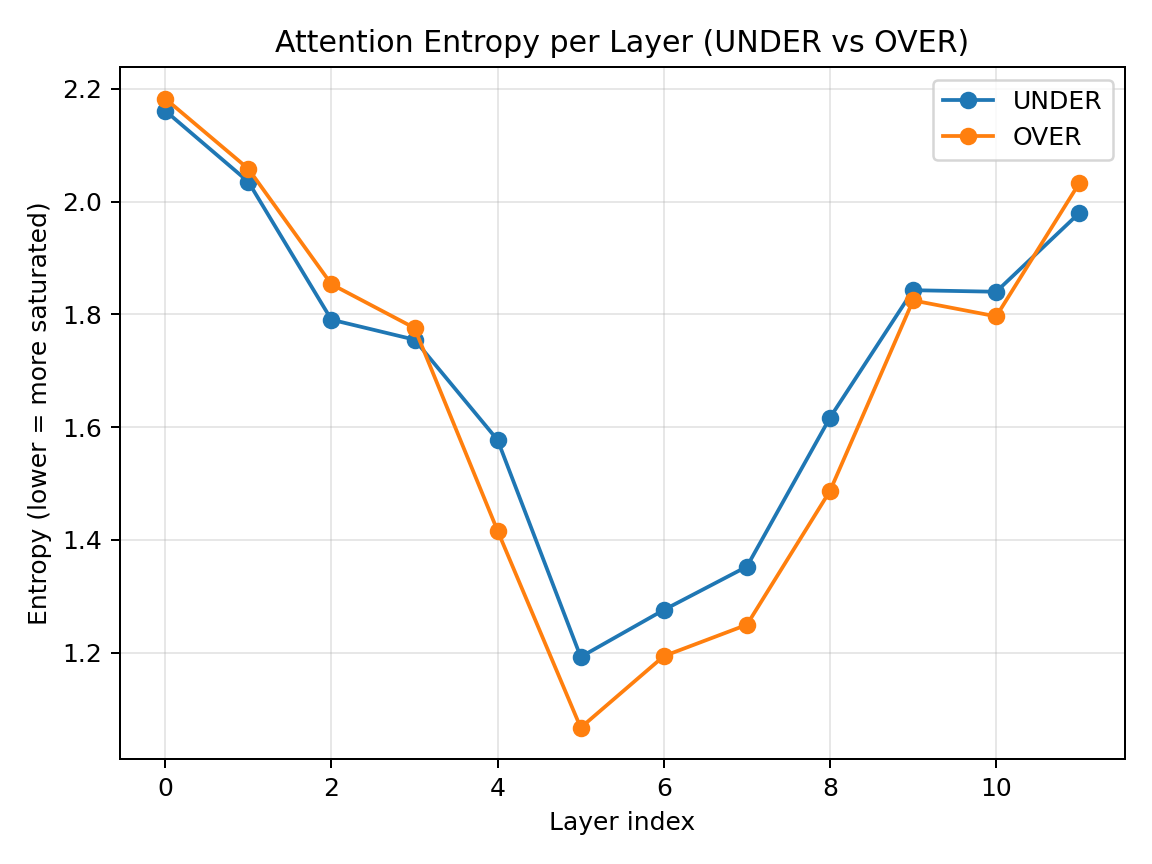}
\caption{Attention Entropy (UNDER vs. OVER)}\end{subfigure}\hfill
\begin{subfigure}{.48\linewidth}\centering
\safeincludegraphics[width=\linewidth]{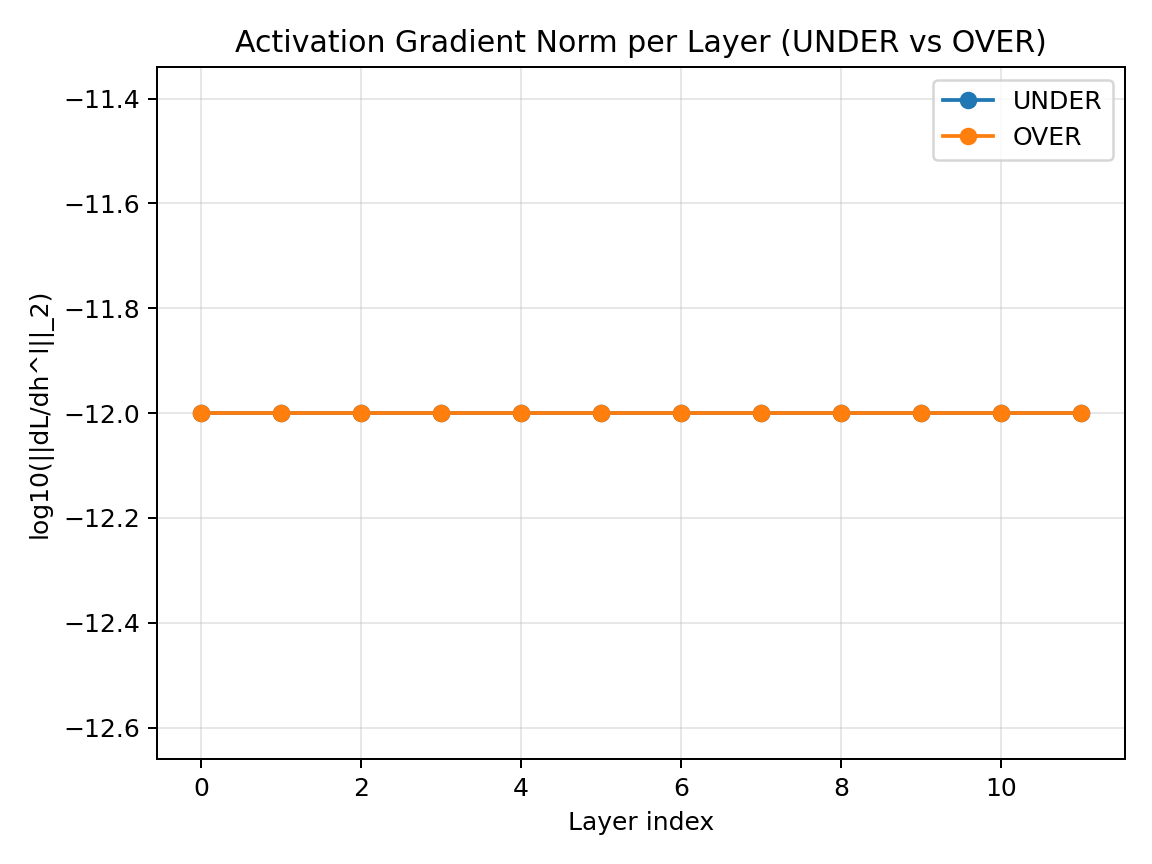}
\caption{Activation Gradient Norm (UNDER vs. OVER)}\end{subfigure}

\vspace{0.3em}
\begin{subfigure}{.48\linewidth}\centering
\safeincludegraphics[width=\linewidth]{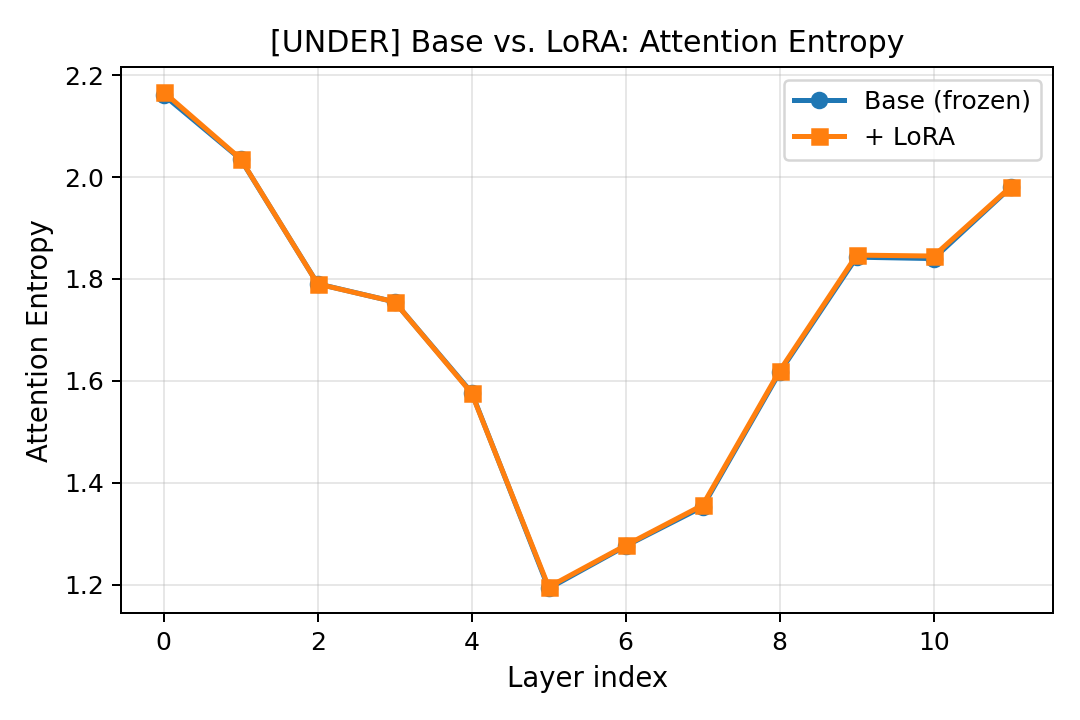}
\caption{UNDER Model: Base vs. LoRA Entropy}\end{subfigure}\hfill
\begin{subfigure}{.48\linewidth}\centering
\safeincludegraphics[width=\linewidth]{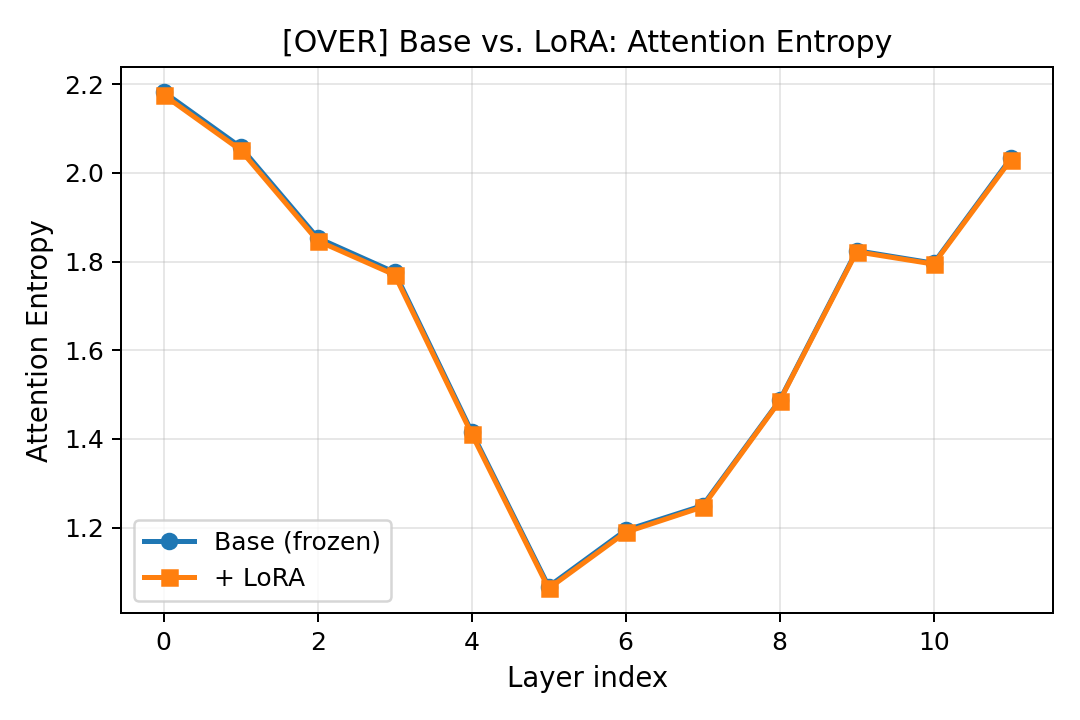}
\caption{OVER Model: Base vs. LoRA Entropy}\end{subfigure}
\caption{Selective LoRA injection: OVER benefits from inflection-layer LoRA while UNDER shows degradation, demonstrating that \textbf{gradient suppression confines models to high-level composition; enabling low-level reconstruction requires full pathway unblocking}.}
\label{fig:lora_compare}
\end{figure}

We apply our SKI metric (Section~5) to automatically identify inflection layers. The algorithm identifies layers $\{0,1,4,5,6\}$ for LoRA injection (rank 4, $\alpha\!=\!16$). Figure~\ref{fig:lora_compare} shows attention entropy and activation gradients for LoRA-based fine-tuning. Key observations:

\begin{itemize}[leftmargin=1.2em]
\item \textbf{Selective beats uniform:} Selective LoRA (91.59$\pm$0.15\%) outperforms LoRA Everywhere (91.46$\pm$0.23\%) despite using 3$\times$ fewer parameters (0.3M vs.\ 0.9M). This validates our diagnostic-driven strategy: \emph{injecting at inflection layers is more effective than uniform application}.
\item \textbf{OVER+LoRA:} Achieves the highest accuracy across all methods with 99.7\% fewer parameters than full unfreezing. The injected low-rank pathways in inflection layers (notably Layer 5 with lowest entropy) successfully restore backward flow without requiring full layer training. This demonstrates that \emph{when strong low-level features are already present}, unblocking inflection layers enables upper layers to adapt via high-level composition—the model needs only \emph{pathway restoration}, not feature reconstruction.
\item \textbf{UNDER+LoRA:} Shows slight degradation (90.96$\pm$0.24\% vs.\ 90.81$\pm$0.23\% shallow baseline), revealing the fundamental limitation of selective intervention: when base features are weak, \emph{unblocking gradients alone is insufficient}. The target task requires \emph{low-level feature reconstruction}, which demands full gradient penetration from output to embedding layers—a structural capability that low-rank adapters at inflection layers cannot provide. This confirms that \textbf{gradient suppression confines adaptation to high-level composition; low-level reconstruction requires unblocking the entire pathway}.
\item \textbf{Layer selection consistency:} Both UNDER and OVER identify the same inflection-layer band via SKI across all three random seeds, suggesting that the \emph{structural bottleneck} (attention saturation + gradient cliff) is architecture-driven rather than purely training-regime-dependent.
\end{itemize}

\subsection{Representation Change and Task Separability}
\vspace{-0.2em}

\begin{figure}[H]\centering
\begin{subfigure}{.48\linewidth}\centering
\safeincludegraphics[width=\linewidth]{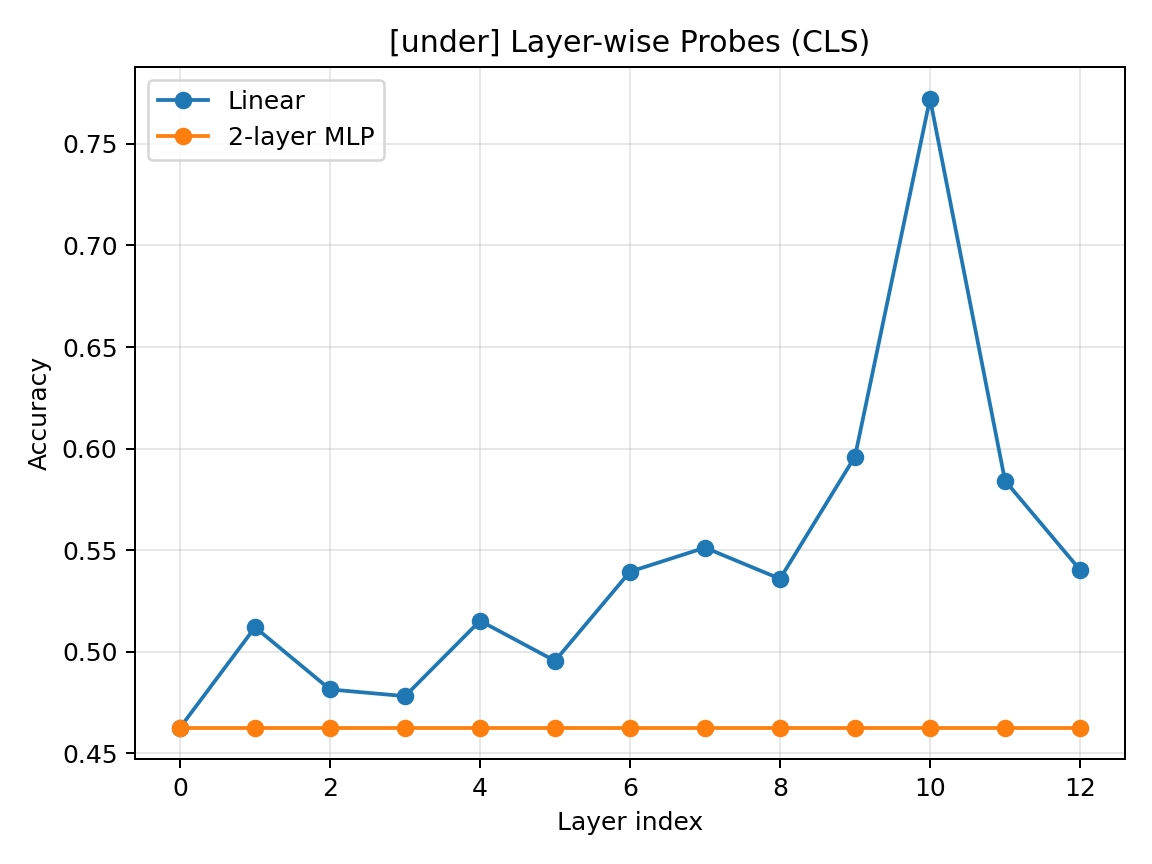}
\caption{Probe Accuracy (UNDER, Shallow Unfreeze)}\end{subfigure}\hfill
\begin{subfigure}{.48\linewidth}\centering
\safeincludegraphics[width=\linewidth]{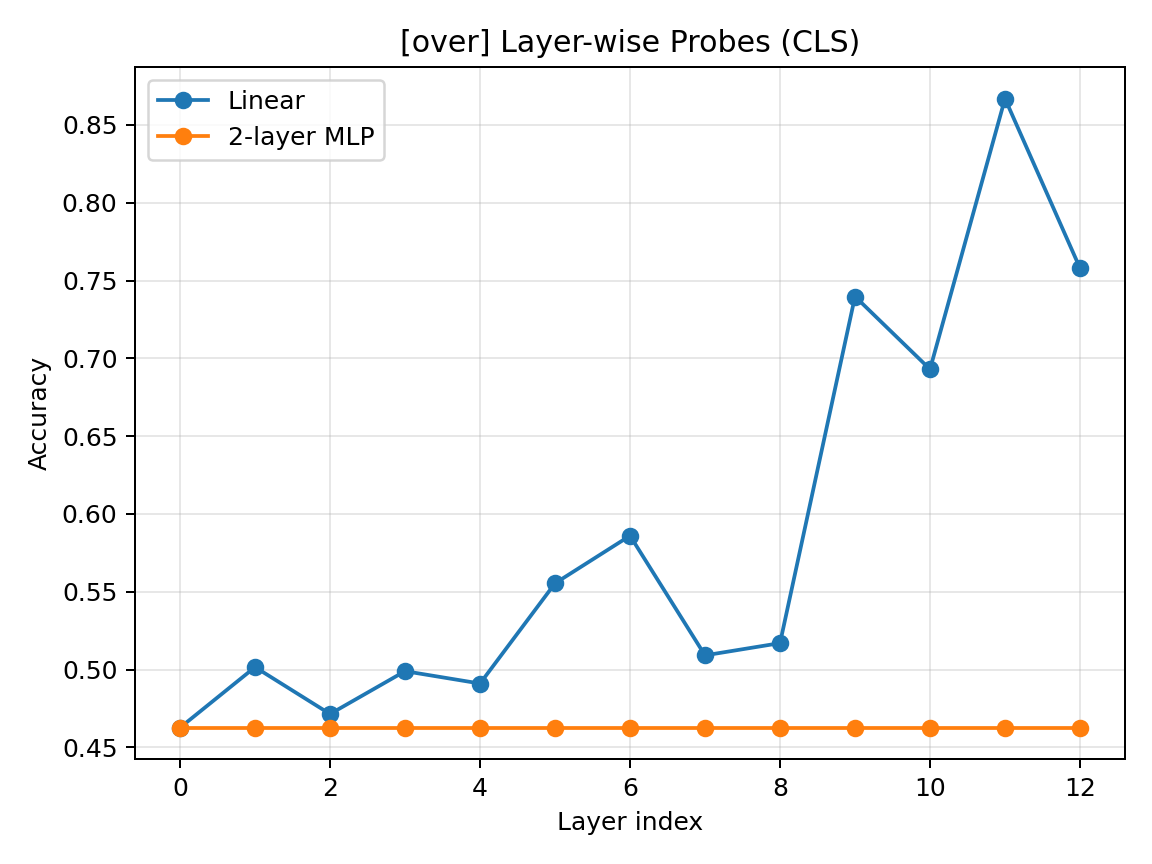}
\caption{Probe Accuracy (OVER, Shallow Unfreeze)}\end{subfigure}

\vspace{0.3em}
\begin{subfigure}{.48\linewidth}\centering
\safeincludegraphics[width=\linewidth]{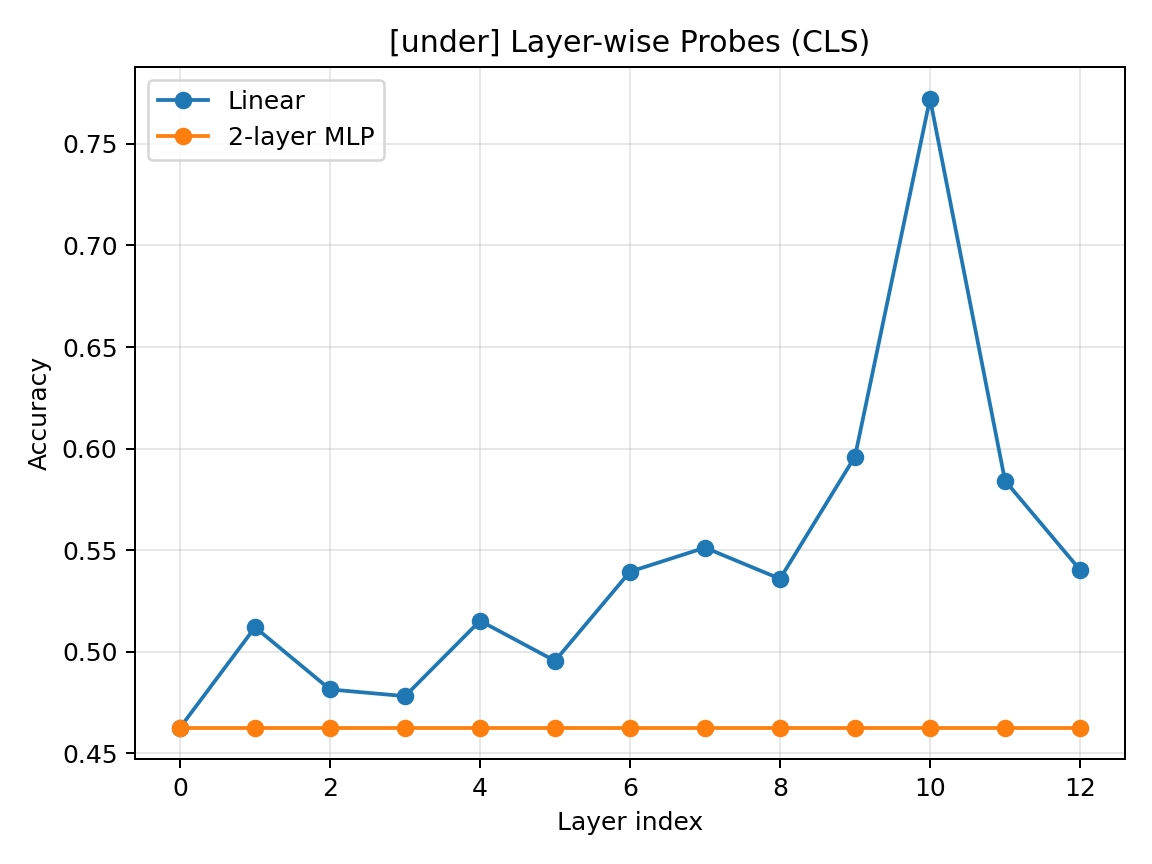}
\caption{Probe Accuracy (UNDER, with LoRA)}\end{subfigure}\hfill
\begin{subfigure}{.48\linewidth}\centering
\safeincludegraphics[width=\linewidth]{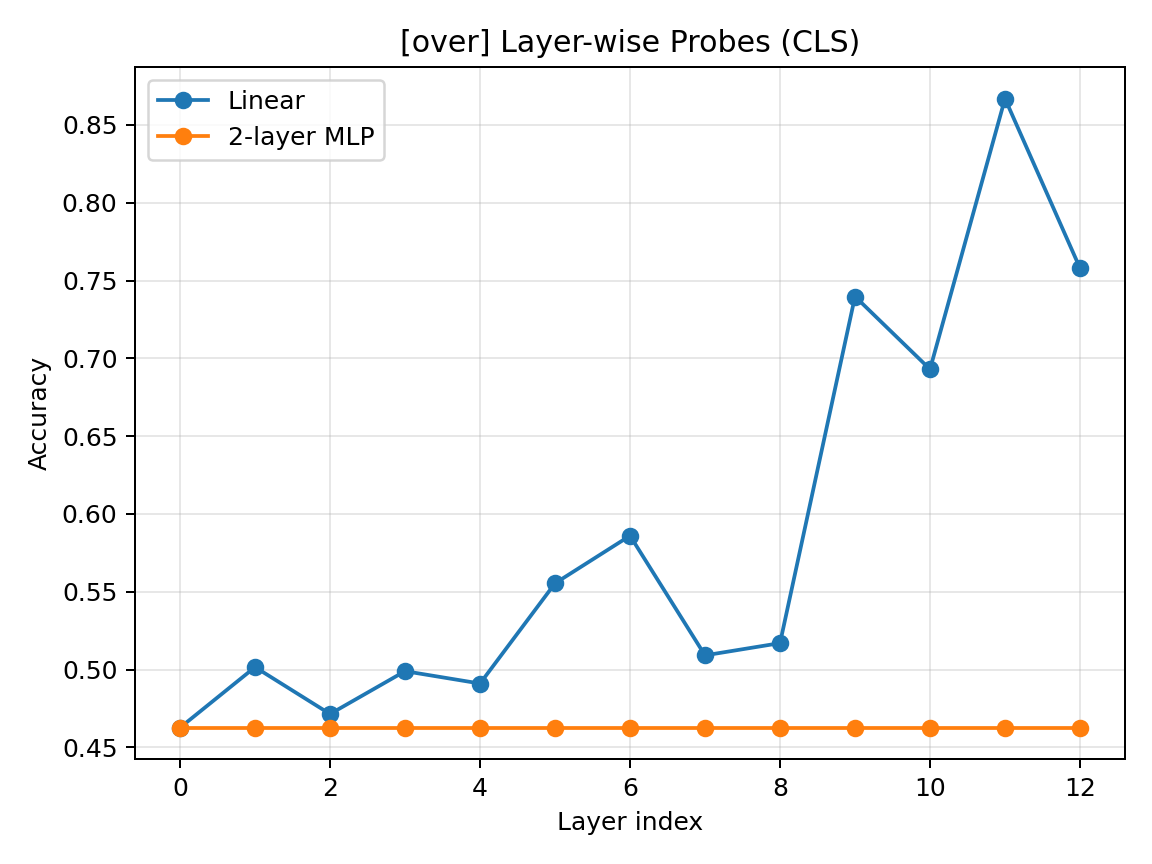}
\caption{Probe Accuracy (OVER, with LoRA)}\end{subfigure}
\caption{Layer-wise probe accuracy: task separability concentrates in upper layers across all settings.}
\label{fig:probes}
\end{figure}

Figure~\ref{fig:probes} shows layer-wise Linear and MLP probe accuracy on [CLS] representations. Task separability \emph{primarily improves in upper layers}, but with a critical difference: UNDER achieves peak probe accuracy at Layer 10 (77.2\%), while OVER peaks at Layer 11 (86.7\%)—a 9.5\% gap. This indicates that (\emph{i}) OVER builds stronger task-relevant representations in the final layer, consistent with over-confident source-domain training; (\emph{ii}) the representation quality gap emerges in the \emph{uppermost} layers, not at inflection layers (Layer 5). In full-unfreezing experiments, $\Delta$CKA measurements show that representation change is most pronounced in upper layers (0.02--0.04 for OVER, 0.03--0.04 for UNDER), while shallow unfreezing exhibits minimal change in frozen layers (near 0) and modest change only in the trainable top-2 layers ($\sim$0.005). Notably, MLP probes show flat accuracy ($\sim$46.3

\vspace{-0.4em}
\section{Discussion}
\vspace{-0.2em}
\textbf{Quantitative gradient suppression:} Our diagnostics reveal stark differences between UNDER and OVER. In shallow unfreezing, OVER exhibits activation gradients $\sim$20$\times$ smaller than UNDER (mean: $1.9\!\times\!10^{-6}$ vs.\ $3.5\!\times\!10^{-5}$), while in full unfreezing, the gap narrows to $\sim$20$\times$ (mean: $3.1\!\times\!10^{-5}$ vs.\ $6.5\!\times\!10^{-4}$). Parameter gradients also differ: OVER's mean parameter gradient is $\sim$6$\times$ smaller in shallow unfreezing (0.24 vs.\ 1.50) but comparable in full unfreezing (1.32 vs.\ 1.67), suggesting that \emph{gradient suppression is most severe when high layers alone must compensate}.

\textbf{High-level composition is the default adaptation mode:} When inflection layers exhibit gradient suppression, models are structurally constrained to adapt in upper layers only—effectively solving new tasks by \emph{recombining existing high-level representations}. OVER models have strong base features locked in lower layers: small interventions (LoRA, $\sim$0.3M params) at inflection layers can \emph{unblock} backward flow, enabling upper layers to compose these features for the target task—OVER+LoRA achieves 91.59$\pm$0.15\% accuracy, the highest across all methods. In contrast, UNDER models lack strong base features: unblocking gradient pathways alone is insufficient without comprehensive \emph{low-level reconstruction}, which requires full gradient penetration and larger structural freedom—UNDER+LoRA achieves only 90.96$\pm$0.24\%. This asymmetry reveals that \textbf{gradient suppression forces high-level composition; overcoming it requires enabling low-level reconstruction}.  

\textbf{Selective injection outperforms uniform LoRA:} The LoRA Everywhere baseline (91.46$\pm$0.23\%) demonstrates that simply adding low-rank pathways to all layers does not improve over shallow unfreezing, despite using 0.9M parameters. In contrast, selective injection at inflection layers achieves higher accuracy (91.59$\pm$0.15\%) with 3$\times$ fewer parameters (0.3M). This validates our core hypothesis: \emph{diagnostic-driven intervention is more effective than blind uniform application}.  

\textbf{Layer positioning is critical:} Injecting adapters uniformly across all layers is not robust; \emph{diagnose-first} narrow-band injection (here, layers $\{0,1,4,5,6\}$ centered on the entropy minimum at Layer 5) excels in stability, sample efficiency, and interpretability. The consistency of inflection-layer identification across UNDER/OVER suggests architectural universality.  

\textbf{Why not inject into upper layers?} One might expect upper layers (9--11), which show highest probe accuracy and largest $\Delta$CKA in baseline experiments, to benefit most from LoRA. However, these layers are \emph{already} being updated effectively (high parameter gradients, no gradient cliff). Our hypothesis is that the bottleneck lies in \emph{middle layers} where saturation \emph{blocks} information flow; once unblocked, upper layers can adapt naturally. This is supported by OVER+LoRA maintaining performance despite not injecting into layers 7--11.

\vspace{-0.4em}
\section{Limitations}
\vspace{-0.2em}
(\emph{i}) \textbf{Attention entropy as proxy:} Entropy is a \emph{correlational proxy} for saturation, not a causal mechanism; interventional studies (e.g., temperature-scaled attention) are needed to establish causality.  
(\emph{ii}) \textbf{Limited scope:} Results hold under BERT-base on English sentiment/review transfer—cross-lingual, cross-domain (e.g., NER, QA), and larger-scale models (RoBERTa, GPT-style decoders, LLaMA) require separate validation.  
(\emph{iii}) \textbf{Heuristic layer selection:} SKI is heuristic with greedy implementation; learnable weighting $\alpha$, multi-metric fusion, or gradient-based layer-importance scoring may improve robustness.  
(\emph{iv}) \textbf{Lack of PEFT baselines:} Beyond LoRA Everywhere, comparisons with Adapters, Prefix-tuning, BitFit, and IA$^3$ would strengthen claims about ``diagnose-first'' strategies.

\vspace{-0.4em}
\section{Future Directions}
\vspace{-0.2em}
\textbf{Two-stage ``debiasing--relearning'':} Before target-domain fine-tuning, introduce a short \emph{debiasing} phase (e.g., increasing attention temperature, maximizing source-class logit entropy, or mild gradient ascent on source patterns), then switch to standard/LoRA fine-tuning. We expect to observe \emph{validation loss rise-then-fall} and synchronized recovery of activation gradients and $\Delta$CKA at inflection layers.  
\textbf{Zero/near-zero-parameter plasticity injection:} At entropy-valley layers, perform \emph{head dropout/re-initialization}, \emph{selective FFN layer reset}, or apply \emph{attention temperature annealing}~$(T:1.5\!\rightarrow\!1.0)$ and \emph{entropy regularization} (minimal weight, short-duration) during early training. These operations can be directly evaluated within our metric framework.  
\textbf{Controlled pattern-rebuilding validation:} Construct \emph{pattern-specific test subsets} (e.g., semantically distinct triggers in target domain), and monitor whether low/middle-layer $\Delta$CKA and MLP probe accuracy significantly improve, validating whether ``new pattern rebuilding'' truly occurs.  
\textbf{Extension to other modalities and architectures:} Apply the diagnostic framework to vision Transformers (ViT), multimodal models (CLIP, Flamingo), and sequence-to-sequence models (T5, BART), examining whether inflection layers exhibit similar saturation--gradient coupling.

\vspace{-0.4em}
\section{Conclusion}
\vspace{-0.2em}
We ground the intuition of ``output saturation $\Rightarrow$ gradient suppression'' in layer-wise observable metrics, revealing a fundamental mechanism: \emph{gradient suppression at inflection layers confines models to high-level composition of existing features, blocking low-level reconstruction}. We propose selective LoRA injection based on \emph{inflection layers} to restore suppressed backward pathways. Experiments demonstrate: when strong base features exist (OVER), unblocking inflection layers enables effective high-level adaptation with minimal parameters; when base features are weak (UNDER), low-level reconstruction requires full gradient penetration beyond what selective adapters can provide. This explains why pre-trained models excel at similar tasks (composition suffices) but struggle when target domains demand fundamentally different abstractions (reconstruction required). We envision this ``diagnose--intervene'' pipeline as a general-purpose tool for transfer learning practitioners, enabling \emph{measurable, actionable, and reproducible} adaptation strategies across diverse domains.

\vspace{-0.4em}
\paragraph{Reproducibility}
All experiments are repeated across three random seeds (42, 43, 44) with results reported as mean$\pm$std. Detailed experimental configurations and hyperparameters are provided in Section~6. Code and data will be made available upon acceptance.

\end{document}